\documentclass{article}
\usepackage{blindtext}
\usepackage[utf8]{inputenc}
\usepackage{graphicx}

\title{Job Detection in Twitter  }
\author{Besat Kassaie}

\date{Spring 2016}

\begin{document}

\maketitle

\begin{abstract}
In this report, we propose a new application for twitter data called \textit{job detection}. We identify people's job category based on their tweets. As a preliminary work, we limiteour task to identify only IT workers from other job holders. We have used and compared both simple bag of words model and a document representation based on Skip-gram model. Our results show that the model based on Skip-gram, achieves a 76\% precision and 82\% recall.
\footnote{This document is the project report prepared for CS 886, University of Waterloo}
\end{abstract}

\section{Introduction}

Internet users are producing large amount of data with almost no cost for people and companies who can exploit this data for their own benefits. Besides, more sophisticated data analysis techniques are available nowadays comparing to even one decade ago. Using data analysis techniques invaluable information can be induced from abundant data available on the web. 

Social networks, such as twitter, are one of the most popular class of applications gathering a lot of information in different formats such as text, image, and video. So far people have worked on twitter data from interesting and different aspects. They could extract highly accurate information in terms of sentiments \cite{twitter_sentiment}, fine grain location information \cite{Li_2014}, churn prediction \cite{DBLP_hadi} , topic detection  \cite{Spina_2014} ,and so on. 

Our main contribution in this work is to detect twitter users' job based on the textual content of their tweets. We did not find any similar work to target this application on twitter. Job detection from tweets have many potential applications such as targeted commercial advertisement, credit scoring ,and so on. 

We faced some challenges in this project. First, there is a huge diversity among career domains, from cosmetic services to medical fields and from astronauts to miners. We would need a huge set of samples from twitter users to cover all of these categories. On the other hand, it is not feasible to crawl such dataset in a short time, considering the limitations which twitter imposes on the rate of fetching tweets and also our hardware resources. The other challenge is that we need to label each user with a job category in our dataset. This is not easy task to produce such a training data to cover all job categories. Also, not all people reveal their information in twitter due to their career category. For example it could be assumed that journalists are more likely to have an active twitter account than miners. 

Considering all those challenges, we limited the target job categories into Information Technology related jobs. By this assumption, we need to gather much less data than what is needed to cover all job categories. Also we can assume that many of IT workers are likely to use twitter. Finally as we are familiar with job titles in this field, we could label data easily and accurately.

Another contribution of this work is applying Skip-gram model for computing word vectors and using KMeans for representing documents by word vectors. We showed that our model based on word vectors outperforms simple bag of word representation of documents. 

In the next sections we first introduce our data gathering strategies and methods. Then we present the implementation details and results and finally we give the conclusions and future works.

\section{Method and Data}
For this work we needed a set of twitter users labeled with their job. As there is no such dataset we had to create our own dataset. Compiling such dataset for all jobs would take a lot of time and resources. So in this preliminary work we focused on detecting people with IT jobs. So our labeled dataset would include people labeled as IT workers or non IT workers. 

In this work we investigate two architectures for classification based on different approaches for document representation. The first one relies on the well-known bag of words model for document representation. For the second model, we use a document representation based on the term vectors which are extracted by word2vec. Word2vec is a tool for computing vector representations of words introduced by a team of researchers at Google. We explain more about word2vec in the next section.

We also propose more in detail explanation for data gathering strategies as well as our document representation and classification techniques in next sections.
\subsection{Word2Vec}

Although representing words as indices in the dataset vocabulary for using in NLP tasks has many advantages such as simplicity and fast model creation, it ignores possible and obvious similarities between words. For example the simple techniques of word representation cannot detect the semantic similarity between ‘King’ and ‘Man’ as well as the syntactic similarities between ‘Flowers’ and ‘Cats’. The ideal word representation for many applications is a representation which is able to detect all possible similarities and also preserve regularities between vectors as much as possible. The regularities are observed as constant vector offsets between pairs of words sharing a particular relationship \cite{export_189726}. Some examples of these regularities are listed below: 

vector ("King") - vector ("Man") + vector ("Woman") = vector ("Queen")

vector("apple")- vector("apples")= vector("car") - vector("cars)

\begin{figure}[h!]
	\label{fig:skp}
	\centering
	\includegraphics[width=1\textwidth]{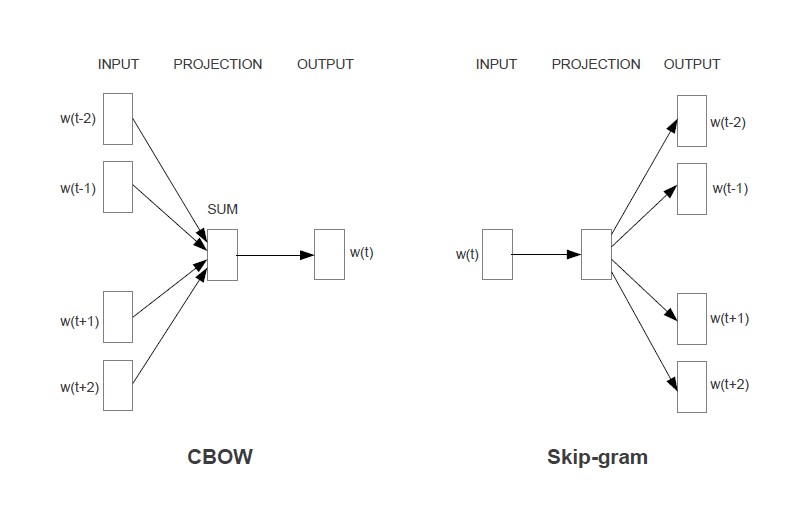}
	\caption{The CBOW architecture predicts the current word based on the context, and the Skip-gram predicts surrounding words given the current word \cite{export_189726}}
\end{figure}

In order to capture these regularities there are different models such as Bag-of-Words Model (known as continuous bag of words, or CBOW) and Skip-gram Model (Figure \ref{fig:skp}) . The first model uses context to predict a target word and the second model uses a word to predict a target context.

We use the Skip-gram method because it produces more accurate results on large datasets. In this paper we used Word2Vec \cite{_word2vec_}  as the tool for computing vector representations of words which uses the Skip-gram model.

In the Skip-gram model each current word is used as an input to a log-linear classifier with continuous projection layer to predict words within a certain range before and after the current word. Apparently a larger range results in better word vectors and at the same time imposes more computational costs. As the distance of context words from the current word is increased they get less related to the current word so Skip-gram model assigns less weight to the distant words. To do so the Skip-gram model samples less from distant words in the training dataset. 

The Skip-gram model trains high dimensional word vectors on a large dataset and detects very accurate semantic relationships between words which can be applied on NLP applications and results in surprising results.

\subsection{Data Preparation}

We used two techniques for building our dataset. As our first approach we tried to use Linkedin profiles for obtaining some auto labeled users. In this approach we look for people in Linkedin who indicated that they are working in some IT related jobs. Then we could use their name and try to find an associated twitter ID to their LinkedIn name. 

There are some challenges in this approach. First of all there are no dictionary of IT job titles. To deal with this issue we compiled a set of job names to cover such jobs. This dictionary includes 183 job titles. Next challenge is that there is no free API in Linkedin for searching over people. To deal with this challenge we used the free Bing search engine API. We used a search query like: \textit{ \{jobtitle\} + 'site:ca.linkedin.com/in'} for obtaining name of people in Linkedin who mentioned one of our job titles in the description of their profile. We gathered about 4092 of such Linkedin users for our IT job titles. The final challenge is to match this names with twitter Ids. For this part we first used UserSearch API in twitter. Using that API we gathered 43719 candidate twitter Ids. As there are multiple twitter IDs for each Linkedin Id we need to filter the twitter Ids. To do so, in the next step, we got the profile data of the candidate twitter Ids. We used the description of these  profiles and calculate their \textit{Jaccard similarity} scores to the description  of Linkedin profiles. We filtered out the twitter Ids which had a similarity score under 0.5. By this method we gathered a total of 277 twitter Ids with IT related jobs. We gathered the most recent 3200 tweets of this users and included them in our dataset.

The second approach is a manual method for gathering data in which we used the twitter API directly. Here we again used the UserSearch API in twitter and queried using each job title. The gathered data were not clean and many of the twitter Ids were not actually related to people working in IT jobs. We used a manual process of verifying each of this candidate twitter Ids based on the description of their twitter profiles. By using both previous approaches we gathered a total of 805 positive and 574 negative samples. Like the previous method we fetched the most recent 3200 tweets per user for both set of users.

In our work we also needed some unlabeled data for a pretraining phase. We gathered the last 3200 tweets of for each of about 7237 random twitter users for this purpose.

\subsection{Classification Architectures}

In our work we do not look for the signals in a single tweet level, we combine the user's most recent tweets obtained from their time lines to create a large document. In our first approach, we simply extracted the document representations of the labeled dataset based on occurrence of 5000 terms as features. Then a Naive Bayes model is used for classification of these documents. This results in some high dimensional and sparse representation for documents.

In the second approach we use a more succinct representation for documents. In this approach first we train word2vec model over our set of unlabeled user time lines. As described before, this is an unsupervised model that can provide some vectorized representation for each word which is also semantically meaningful. Having this word representations we build a proper representation for documents in our labeled dataset.

Different approaches could be considered for building the document representations based on the vectors obtained from word2vec. For example the simplest one is using the bag of words representation and replacing the ones with word2vec vectors and zeros with a zero vectors with a length equal to word2vec vectors. However this method results in very lengthy document vectors. However, considering the few amount of labeled data, this leads to low classification performance. There are other suggested approaches such as using average of the word vectors in a document. We applied this approach to create the feature vector representing each document. Finally, like the bag of words, we use the Naive Bayes model for classifying the extracted document representations.

\section{Implementation and Results}
We used \textit{Python} along with \textit{sklearn} package as our main machine learning package beside gensim package which contains an implementation of word2vec in Python. We conducted two experiments in this work. As our first experiment we used the bag of words model where in the second experiment we used the word vectors obtained from the a pre-training phase by using word2vec method.

For the bag of word model, we represented each document by 5000 terms as features. We used 80 percent of  randomly selected labeled dataset  for training and its remaining as the test set. We applied a Naive Bayes model over the training set for classification. The results are represented in Table \ref{tb:results}.

In the next experiment we used the document representation based on word2vec in our classification. We used word2vec for pretraining and extracting word vectors from a set of timelines of 7237 twitter users. Using the average vectors as described in the previous section we ended up with representing each document by a vector of length 200. Again we used 80 percent of labeled data for training and the remaining for testing.  RandomForest model is used for classification. The results are represented in table \ref{tb:results}.

%

	\begin{table}
		
			\caption{The performance results for the two employed document representations}
			\begin{center}
				\label{tb:results}
	\begin{tabular}{  l | l | l | l }
		
		  & Precision& Recall & F1-Measure\\
		\hline
		Bag of Words & 0.69& 0.79& 0.74\\
		\hline
		Word2Vec& 0.76& 0.82& 0.79\\

	\end{tabular}
\end{center}		
\end{table}

As shown in table \ref{tb:results} pre training and proposed document representation improved all of performance measures.  Beside these improvements we may also note that by using pre training we achieved the higher performance by using a highly dense vector representation of only 200 features.

\section{Conclusion and Future Works}

Here we presented our work on detecting twitter users jobs based on their tweets. This is a new application of tweeter data. Due to the large number of job titles and job categories, in this preliminary work we tried to just recognize people who have an IT related job title. By using the auto labeling method that we described in the data gathering section, we can add more labeled data for other jobs to recognize other job categories as well.

The other contribution in this work is using deep learning to produce document representations  and showing that it can produce better results than bag of words model for  this specific application. We used word2vec to extract word vectors and proposed a new document representation based on these vectors. We have adopted rather a naive approach for combining the word vectors to create a document representation which can be improved. We could concatenate the word vectors then using  dimension reduction techniques we create a fixed length documents that can be fed into classifiers. 

We have started to extend the job categories by extracting more data and labeling them by using our cross checking mechanism between Linkedin and Twitter. We also extend the job title dictionary by including the job titles from Canada NoC. This includes about 40000 job categories and job titles.

Another path for extending this work which we are pursuing is using Convolution Neural Networks over the word vectors word2vec. Based on \cite{KalchbrennerACL2014,kim_convolutional} we think that using this approach may produce superior results.

\bibliographystyle{acm}
\bibliography{ref}

\end{document}